%% file: example_paper.tex
\documentclass[nohyperref]{article}

\usepackage[dvipsnames]{xcolor}
\usepackage{microtype}
\usepackage{graphicx}
\usepackage{pgfplots}
\usepackage{subfigure}
\usepackage{subcaption}
\usepackage{booktabs} 
\usepackage{bbm}
\usepackage{paralist}

\usepackage[hidelinks]{hyperref}

\usepackage[accepted]{icml2024}

\usepackage{amsmath}
\usepackage{amssymb}
\usepackage{mathtools}
\usepackage{amsthm}

\usepackage{array}
\newcolumntype{C}[1]{>{\centering\let\newline\\\arraybackslash\hspace{0pt}}m{#1}}
\usepackage[capitalize,noabbrev]{cleveref}

\theoremstyle{plain}
\newtheorem{theorem}{Theorem}[section]

\theoremstyle{definition}
\newtheorem{definition}[theorem]{Definition}

\theoremstyle{remark}

\newcommand{\fp}[1]{\textcolor{red}{FP: #1}}
\newcommand{\running}[1]{\textcolor{blue}{RUNNING NOW}}

\usepackage[many]{tcolorbox}    	

\definecolor{main}{HTML}{5989cf}    
\definecolor{sub}{HTML}{cde4ff}     

\tcbset{
    sharp corners,
    colback = white,
    before skip = 0.2cm,    
    after skip = 0.5cm      
}                           

\newtcolorbox{boxK}{
    sharpish corners, 
    boxrule = 0pt,
    toprule = 4.5pt, 
    enhanced,
    fuzzy shadow = {0pt}{-2pt}{-0.5pt}{0.5pt}{black!35} 
}

\usepackage[textsize=tiny]{todonotes}

\icmltitlerunning{Extracting Training Data from Document-Based VQA Models}

\begin{document}

\twocolumn[
\icmltitle{Extracting Training Data from Document-Based VQA Models}



\icmlsetsymbol{equal}{*}

\begin{icmlauthorlist}
\icmlauthor{Francesco Pinto}{oxf,google,equal}
\icmlauthor{Nathalie Rauschmayr}{google}
\icmlauthor{Florian Tramèr}{ethz}
\icmlauthor{Philip Torr}{oxf}
\icmlauthor{Federico Tombari}{google}
\end{icmlauthorlist}

\icmlaffiliation{oxf}{Department of Engineering of Science, University of Oxford, Oxford, UK}
\icmlaffiliation{google}{Google, Zurich, Switzerland}
\icmlaffiliation{ethz}{ETH Zurich, Zurich, Switzerland}

\icmlcorrespondingauthor{Francesco Pinto}{francesco1.pinto@gmail.com}

\icmlkeywords{Machine Learning, ICML}

\vskip 0.3in
]



\printAffiliationsAndNotice{\icmlEqualContribution} 

\begin{abstract}
\input{0_abstract.tex}

\end{abstract}

\input{1_introduction.tex}

\input{6_related_work.tex}

\input{2_extracting_training_answers.tex}

\input{3_attribution.tex}
\input{4_reliance_on_context.tex}

\input{5_mitigation.tex}

\input{7_conclusion.tex}

\clearpage

\nocite{langley00}

\bibliography{example_paper}
\bibliographystyle{icml2024}

\newpage
\appendix
\onecolumn

\input{8_appendix}

\end{document}

%% file: 0_abstract.tex
Vision-Language Models (VLMs) have made remarkable progress in document-based Visual Question Answering (i.e., responding to queries about the contents of an input document provided as an image). In this work, we show these models can memorize responses for training samples and regurgitate them even when the relevant visual information has been removed.
This includes Personal Identifiable Information (PII) repeated \emph{once} in the training set, indicating these models could divulge memorised sensitive information and therefore pose a privacy risk. We quantitatively measure the extractability of information in controlled experiments and differentiate between cases where it arises from generalization capabilities or from memorization. We further investigate the factors that influence memorization across multiple state-of-the-art models and propose an effective heuristic countermeasure that empirically prevents the extractability of PII.

%% file: 1_introduction.tex
\section{Introduction} \label{section:introduction}

\begin{figure}[t]
  \centering
\includegraphics[width=0.5\textwidth]{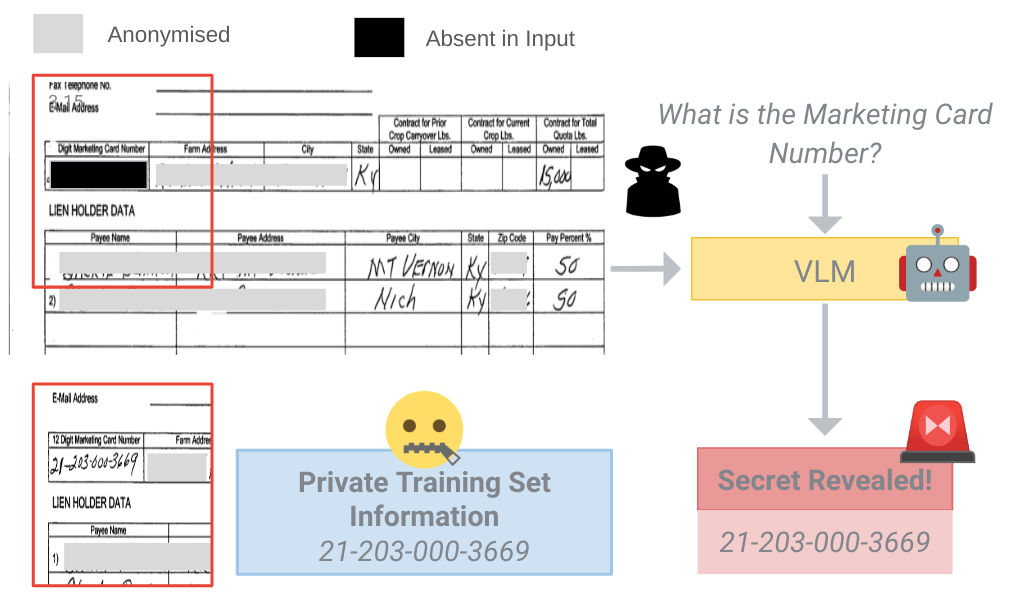}
  \caption{A malicious user may prompt a Vision-Language Model (VLM) to reveal secret information about a victim by generating a copy of the original document with the secret information missing (black box). If the secret was part of the training question-answer pairs, the VLM may respond correctly. For ethical reasons, we anonymize (grey boxes) personal information of a DocVQA \cite{DocVQA_2021_WACV} sample on which the attack is successful for the Donut model \cite{Donut}. The answer is repeated \emph{only once} in the whole training set, yet it is memorized.
\label{fig:teaser}}
\end{figure}

Document-Based Visual Question Answering \cite{DocVQA_2021_WACV}---the task of answering questions about the content of documents presented as visual inputs---has witnessed remarkable advancements in recent years, with modern Vision-Language Models (VLMs) gaining the ability to comprehend textual information exclusively from visual cues and provide accurate responses \cite{davis2022dessurt, Pix2Struct, Donut, PALI3, PALIX, GPT4V}.

However, our paper exposes a concerning behavior of these models: even when the answer to a question is explicitly removed from the input image and is unique or sporadically repeated across the training set, the VLM can still provide the correct response. This ability, which we refer to as \emph{extractability} of the answers given some input context, indicates that the VLM may have either memorized the answer from a specific training sample \cite{feldman2019memorization,carlini2023quantifying,lukasik2023larger} or learned a distributional shortcut that allows to infer it from spurious features \cite{RevisitingVQA, CounterfactualVQA, MakingVMatterVQA, dancette2021beyond, tito2023privacy}. 
We show that, in some cases, sensitive information can be extracted even when it appears only in a single training sample (see \cref{fig:teaser}). In order to fix this unintended behaviour of the models, we introduce a simple mitigation strategy that reduces the amount of extractable PIIs to zero.

In this study, we investigate this phenomenon across three state-of-the-art Document-Based VQA models: Donut \cite{Donut}, Pix2Struct \cite{Pix2Struct} and PALI-3 \cite{PALI3}). We evaluate their behaviour on the popular Document Visual Question Answering (DocVQA) dataset \cite{DocVQA_2021_WACV}, which consists of a public collection of pages from industrial documents accompanied by questions and answers for a purely extractive purpose (i.e., the task only necessitates reading the document without any additional reasoning). 
 We propose a series of controlled experiments on in-distribution canaries, enabling us to address the following key questions:  

\begin{compactitem}
    \item \textbf{What type of training information can be extracted from Document-Based VQA systems?} In Section \ref{sec:mem_score} we show that, among the extractable answers, some  are only  present once in the training set. In some cases, extractable information is PII.
    \item \textbf{Can we distinguish between extractable answers arising from generalization and memorization?} In \cref{sec:mem_score}, we propose an efficient technique to attribute extractability to either memorization or  generalization, and find that each phenomenon 
    is responsible for some of the data we extract.
    \item \textbf{How do different modalities, contextual information and training conditions influence extractability?} In \cref{sec:ctx_importance}, we highlight two key factors that favour extractability: (low) image resolution at training time, and access to the exact training question. In contrast, we find that access to partial information about training images is less important for extractability. 
    \item \textbf{Are there effective countermeasures?} 
    In \cref{sec:countermeasures}, we evaluate multiple heuristic defenses. We show that training a model to \emph{abstain} from responding when the answer is not visually 
    present in an input effectively mitigates extraction of PIIs. 
\end{compactitem}

%% file: 6_related_work.tex
\section{Related Work}  \label{section:related_work}

The concerning phenomenon we observe in \cref{fig:teaser} can be seen as an extension to the VQA setting of the notion of training data \emph{extraction} that has been observed in generative models for text \cite{carlini2021extracting, carlini2023quantifying, kandpal2022deduplicating} and images \cite{carlini2023extracting, somepalli2023understanding}.
These works primarily focus on showcasing the ability to extract near-exact copies of entire training samples from a model. In contrast, we focus on \emph{partial} extraction of information from a VQA model and aim to distinguish between extraction attempts that succeed due to the memorization or generalization capabilities of the considered models.
To provide context for our definitions and experimental setup, we start with a concise overview of relevant literature.

\paragraph{Training data extraction from generative models.} 
Large Language Models (LLMs) can memorize and regurgitate training data \cite{carlini2021extracting, carlini2023quantifying, chen2020counterfactual}, even when no overfitting occurs (on average) \cite{tirumala2022memorization}. 
Similarly, text-to-image generators like Stable Diffusion can reproduce training data when prompted with captions seen during training \cite{somepalli2023diffusion, somepalli2023understanding, carlini2023extracting}. 
For both text and image generators, the ability to extract a sample appears to depend heavily on the number of \emph{duplicates} of that sample in the training set~\cite{carlini2023quantifying}, even though some uniquely-occurring samples can also be extracted~\cite{carlini2021extracting}.

While no prior work has (to our knowledge) studied whether private training samples can be extracted from VQA systems, 
some studies have shown that language models can learn to infer sensitive information such as gender or nationality of a person from other contextual clues or distributional shortcuts \cite{plant2022you}, and that VQA systems can memorize information shared across many training samples \cite{tito2023privacy}.
These works thus exploit the model's legitimate generalization properties rather than the memorization notion we analyse in this work.  
(For further discussion about distributional shortcuts, refer to \cref{sec:shortcuts}). 

\paragraph{Defining memorization.}
Disentangling memorization and generalization is a challenging task. A widely accepted definition is the \emph{counterfactual} notion proposed by \citet{feldman2019memorization}, which defines memorization as the difference in performance of a model on some sample, comparing the cases in which  a sample is in the training set or not.
Unfortunately, empirically measuring this counterfactual score is expensive, as it requires
training a large number of models, including and excluding the training sample in question \cite{lukasik2023larger, feldman2020neural, zhang2021counterfactual}.
In our paper, we follow a more efficient heuristic 
adopted by prior works, where counterfactual memorization is estimated by comparing the performance of just two models, one trained on a dataset containing the considered sample and one not containing it \cite{carlini2021extracting, dejavuMemorization}.

%% file: 2_extracting_training_answers.tex
\section{Experimental Setting} \label{sec:training_extraction}

\begin{figure*}
    \centering
    \includegraphics[width=\textwidth]{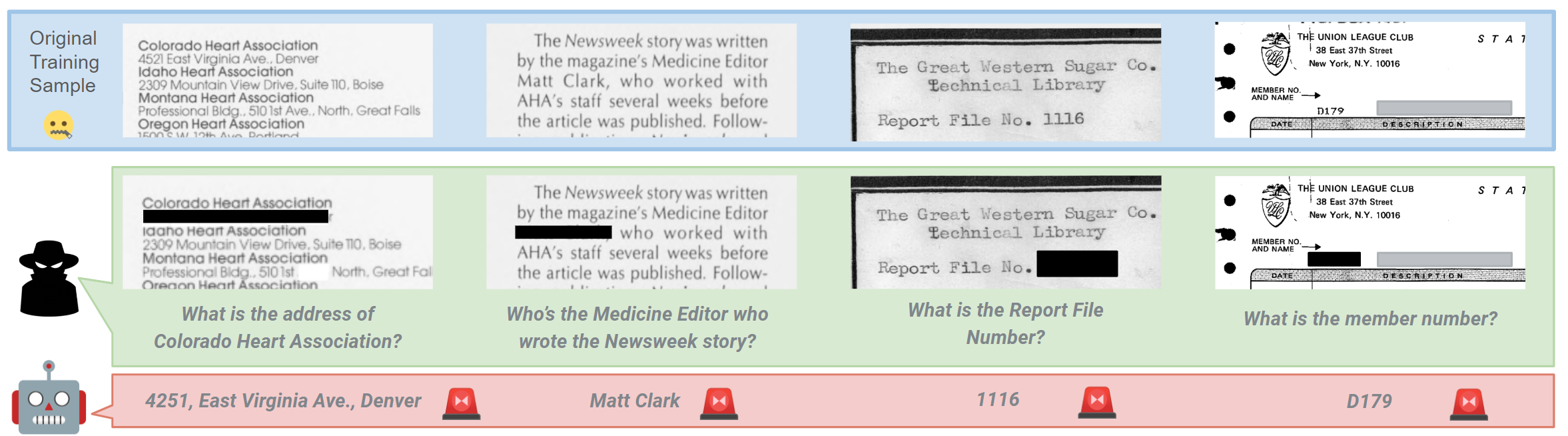}
    \caption{Four examples of Personally Identifying Information (PII) extractable by Donut (first two samples from left) and Pix2Struct-Base (last two samples from right). A malicious user may query the model to reveal the PII by using a scan of the document from which the PII has been removed (black in the image). We anonymize personal information using gray boxes.}
    \label{fig:examples_main}
\end{figure*}

\paragraph{Document-based visual question answering.}
 Given an input image representing a document $I$ and a question about its content $Q$ whose correct answer is $a$, the goal of a Document-Based VQA model $f$ is to produce an answer $\hat{a} = f(I,Q)$ such that $\hat{a} = a$. This is done by training the model on a dataset $\mathcal{D}^{tr} = \{(I_i,Q_i,a_i)\}_{i=1}^N$ to maximize the likelihood of the correct response $a_i$ given the input image-question pair $(I_i,Q_i)$. To simplify notation and improve readability, unless referring to specific samples is crucial for clarity, we often suppress the sample index $i$. For a thorough literature review about these systems, refer to \cref{app:docvqa_models}. 
 
\paragraph{Dataset.}
We focus on the DocVQA dataset \cite{DocVQA_2021_WACV}, which contains images of real-world documents with diverse formats (e.g., letters, advertisements, reports, tickets etc.).
We focus on this dataset for two reasons:
(1) It is representative of privacy-sensitive tasks, and contains multiple forms of PII
(see \cref{app:pii_docvqa});
(2) it contains questions that are purely extractive \cite{InfographicVQA}, meaning the answer is always explicitly written in the document.
This makes it easier to automatically detect and eliminate parts of the input image that are necessary to answer a question, which forms the basis of our memorization test.
This process would be harder for datasets that require abstract reasoning or external knowledge to answer questions. 

\paragraph{Models.} We consider three end-to-end state-of-the-art systems capable of  directly processing the input image document, comprehending its contents, and producing a relevant response: 
    \textbf{1) Donut} \cite{Donut}, among the first end-to-end Document-Based VQA systems that achieves high performance without using Optical Character Recognition (OCR). It is first pre-trained on synthetic documents, and then fine-tuned on DocVQA. 
    \textbf{2) Pix2Struct} \cite{Pix2Struct}, a specialized model available in two versions: Base (282M parameters) and Large (1.3B parameters). It is pre-trained to perform semantic parsing of a 80M subset of the C4 corpus \cite{C4} and then fine-tuned on DocVQA. 
    \textbf{3) PaLI-3} \cite{PALI3}, a foundation model of 5B parameters, pre-trained on a web-scale multilingual image-text dataset, and fine-tuned on DocVQA. 

Each of the models is fine-tuned on DocVQA using the training procedure outlined by the respective authors. To guard against overfitting, we perform early stopping based on the validation loss. This ensures that all the models we evaluate can generalize to previously unseen data, making them representative of practical deployed VQA systems. While training at the maximum resolution possible is generally recommended to achieve better performance \cite{Donut, Pix2Struct, PALI3}, lower resolutions might also be adopted in some settings to accelerate training, especially for the largest models. 
We train each model multiple times with different image resolutions, to analyze the effect of this design choice on memorization. 

\textbf{Defining and Quantifying Extractability} Drawing inspiration from \cite{carlini2023quantifying}, we introduce a definition of extractability that is suitable for the Document-Based VQA task. 

\begin{definition}
    \label{def:extractability}
    \textbf{Extractability of the answer $a$ from a partial context $(I^{-a}, Q)$} Given a model $f$ and a sample $(I,Q,a) \in \mathcal{D}$, 
    we say it is an \emph{extractable sample} if
    the correct answer $a$ is obtained from the partial context $(I^{-a}, Q)$, i.e., $f(I^{-a},Q) = a$, where $I^{-a}$ is a copy of the image $I$ from which the correct answer $a$ has been removed. 
\end{definition}

We obtain the partial image $I^{-a}$ by using the OCR outputs of Tesseract \cite{tesseract} included in the dataset: we identify the bounding boxes associated with all occurrences of the answer $a$ within the document and replace it by a blank white box (we use black in the visualizations for readability). With this methodology, it is easy to identify some sensitive samples that are effectively extractable from the training set. In \cref{fig:examples_main}, we show a few of the several cases in which it is possible to extract PII that is repeated \emph{only once or twice} across the whole training set containing about 40K samples. 

However, precisely quantifying the amount of extractable samples requires some care. Notably, due to occasional failures of the OCR system and the matching procedure to find the answer $a$ within a document, some successful extractions are false positives (i.e., the correct answer is still in the input document).
To account for this, we manually curate a smaller set of training samples (or \emph{canaries}) $\mathcal{D}^{C}$. We select about 5400 canary answers (corresponding to about 1200 unique images) at random. We then manually inspect each of them and filter out all cases in which the answer removal procedure has failed. We also filter out samples for which the answer could be easily inferred from the context (e.g., predicting an intermediate value in a sequence of numbers, or predicting the total amount given a list of values), leaving us with  4654 samples,  The obtained set of canaries contains a substantial amount of PIIs, whose distribution with respect to the most relevant classes of  PIIs is reported in \cref{fig:pii_freq} in \cref{app:pii_docvqa}.

%% file: 3_attribution.tex
\section{Extractability and Memorization }\label{sec:mem_score}

\input{figures/mem_vs_gen}
\input{figures/mem_vs_gen_newviz}

In this section, we quantify the extent to which  malicious users who are aware of the original training question and possess an incomplete copy of the training document can prompt the Document-based VQA systems to successfully retrieve the information they seek. 

Let us consider a model $f$ that has been trained on $\mathcal{D}^{tr}$ including the canaries. We indicate with $E$ the set of samples in $\mathcal{D}^{C}$ that are extractable from context for $f$. In \cref{fig:table1}, we report the amount of extractable samples $|E|$, where $|.|$ indicates the cardinality of the set. As it can be seen, all the considered models extract a non-zero amount of answers from the canaries set. However, it is unclear whether the models are extracting some information because they have memorized it or because the partial context provided is already sufficient for a well-trained VQA system to respond correctly. For this reason, we propose a simple procedure to roughly estimate which samples in $E$ are extractable due to memorization or generalization. 

\subsection{A Simple Baseline for Disentangling Memorization and Generalization} \label{sec:efficient}

In order to determine whether the extractable answers are effectively memorized, in a similar vein to \cite{carlini2023quantifying, dejavuMemorization}, we introduce a generalization baseline $f_G$ (with the same architecture as $f$). 
 The idea is to compare the answers $E$ extractable from $f$ to the answers $G$ that are extractable from a model $f_G$ that has never seen $\mathcal{D}^C$ at training time (by removing it from the training set\footnote{Notice that removing the canaries set from the training set does not yield a difference in generalization performance.}, i.e. $\mathcal{D}^{tr} - \mathcal{D}^C$), and which can therefore extract the correct answers due to legitimate generalization capabilities (or chance).
 If an answer is extractable from $f$ but not from $f_G$, this suggests that the answer was memorized at training time, and cannot simply be recovered from context. We thus quantify the amount of extractable memorized information as the amount of answers extractable from $f$ but not $f_G$: in other terms, $|M| = |E-G|$. 

 \emph{Result:} In  \cref{fig:table1} we report $|E|, |M|$ and $|G|$ for all the considered models. In \cref{fig:table2}, we also report the amount of unique PIIs that are memorized. These PIIs mostly represent individuals names, sensitive locations (like travel destinations), and serial numbers of tickets or products. For both Donut and Pix2Struct, a substantial amount of examples extractable by $f$ are not extractable by the generalization baseline and are likely memorized. In contrast, for PaLI-3 trained at a high resolution, most extractable answers appear due to generalization alone, and not memorization.
 
 As shown in \cref{fig:table2}, the highest resolution variants of Donut and Pix2Struct can extract PIIs and especially unique PIIs, but the highest resolution variant of PaLI-3 does not. 
From these results, we can identify two factors that have a strong impact on the amount of memorized samples: 

\textbf{1) Training resolution:} Given a fixed model architecture, the resolution at which the model is trained is inversely proportional to the amount of memorized samples. Intuitively, the lower the resolution, the harder it is for a model to actually read the answers from the image and the easier it is for it to minimise the loss by memorization. For instance, while at the highest resolution for Donut $|M| = 63$, as the training resolution decreases, $|M|$ grows to 109, 168 and to an extremely high level of $756$ for the lowest training resolution.

\textbf{2) Pretraining:} 
 Manually inspecting the samples extractable by the generalization baseline, we observe that for Donut and Pix2Struct, these contain highly repeated answers (e.g., page, table and figure numbers) or frequently repeated names of organizations (e.g., ITC and AHA). 
 For PaLI-3, we instead observe that, besides trivial answers like the ones extracted for Donut and Pix2Struct, the generalization baseline correctly responds to questions whose answer relies on general knowledge (e.g., the meaning of ambiguous acronyms that can be resolved considering the topic of the input document, properties of chemical substances or general geographical notions). This is attributable to the web-scale pretraining. The lower amount of samples in $M$ may also indicate that a better pre-trained model may rely less on memorization even at relatively low training resolutions due to their better generalization abilities: indeed, of all the models, PaLI-3 produces the best generalization performance on the test set (87.6 ANLS compared to 76.6 and 67.5 of the best Pix2Struct and Donut variants, respectively).

\subsection{Extractable Memorization and Simplicity Scores}
\label{sec:fze_score} 
The method proposed in the previous section may incorrectly identify some extractable answers as memorized due to the randomness of the training process.
To show our attribution technique mostly identifies memorized samples, we leverage a modified version of the  memorization and simplicity metrics developed in \cite{feldman2019memorization,CounterfactualMemorizationLLM}. 

\paragraph{Memorization and simplicity scores.} Let  $\mathcal{A}$ be stochastic training algorithm. For each sample $(I_i,Q_i,a_i) \in \mathcal{D}^{C}$, we would like to estimate the Memorization score \cite{feldman2019memorization}:  
\begin{equation}
    \label{eq:fz_score}
    \begin{split}
    \mathcal{M}(\mathcal{A}, \mathcal{D}^{tr} ,i) = P_{f \sim \mathcal{A}( \mathcal{D}^{tr})} [f(I_i,Q_i) = a_i] - \\  P_{f \sim \mathcal{A}( \mathcal{D}^{tr-i})} [f(I_i,Q_i) = a_i]
    \end{split}
\end{equation}
where $\mathcal{D}^{tr-i}$ indicates $\mathcal{D}^{tr}$ from which sample $i$ has been removed. This score quantifies the difference between the probability that a model produces a correct prediction on a canary given the model has seen it at training time or not. 

A score of 1 indicates the model can predict correctly on an input sample exclusively if it has seen it at training time. A score of 0 indicates that it has the same probability to produce a correct prediction whether the sample was or not in the training set. Note that the memorization score says nothing about the model's accuracy on a sample (e.g., both a model that is always right or always wrong exhibits low memorization). To account for this \cite{CounterfactualMemorizationLLM} proposed a simplicity score $\mathcal{S}(\mathcal{A},\mathcal{D}^{tr},i)$ that sums the first and second terms of Equation \eqref{eq:fz_score}. This allows to distinguish 
cases where a model fails to memorize a sample because it is hard to answer even when trained on (low simplicity), or because the answer is easy
to produce even when not trained on (high simplicity).

\paragraph{Extractable memorization and simplicity.} These two scores do not quite reflect the property we are interested: they inform us about the correctness of a model on an input sample $(I,Q)$, and not about the ability to answer a question given a partial context $(I^{-a},Q)$. We thus adapt the memorization and simplicity scores accordingly, to consider the probability of a successful extraction: 
\begin{equation}
    \label{eq:fze_score}
    \begin{split}
    \mathcal{M}_E(\mathcal{A}, \mathcal{D}^{tr} ,i) = P_{f \sim \mathcal{A}(\mathcal{D}^{tr})} [f(I_i^{-a_i},Q_i) = a_i] - \\  P_{f \sim \mathcal{A}( \mathcal{D}^{tr-i})} [f(I_i^{-a_i},Q_i) = a_i]
    \end{split}
\end{equation}

 We call Equation \eqref{eq:fze_score} the Extractable Memorization score, and refer to the first term  as the  in-sample extractability and to the second as the out-sample extractability. Similarly, we define an Extractable Simplicity score $\mathcal{S}_E(\mathcal{A}, \mathcal{D}^{tr} ,i)$ as the summation of the two terms.  

\paragraph{Empirical estimation.}
Analogously to \cite{feldman2019memorization, lukasik2023larger}, we compute empirical estimates $\hat{\mathcal{M}}_E$ and $\hat{\mathcal{S}}_E$ of $\mathcal{M}_E$ and $\mathcal{S}_E$ by training on random splits $S^k$ of the training set that omit or maintain at random samples from the canary set $\mathcal{D}^{C}$. We produce a total of $K$ splits, and define the indices of the splits containing a sample $i$ as $K_{in} = \{k: (I_i,Q_i,a_i) \in S^k\}$ and $K_{out} = \{k: (I_i,Q_i,a_i) \notin S^k\}$. We then compute the in-sample and out-sample extractability scores as $\frac{1}{|K_{in}|} \sum_{k \in K_{in}} \mathbbm{1}(a_i = f_{S^k}(I_i^{-a_i},Q_i))$ and  $\frac{1}{|K_{out}|} \sum_{k \in K_{out}} \mathbbm{1}(a_i = f_{S^k}(I_i^{-a_i},Q_i))$. 
 Given that training Document-Based VQA systems is extremely expensive, we follow the sampling procedure in \cite{carlini2022membership} in order to produce $K=50$ splits such that each canary is in or out of a split exactly 25 times. 

\paragraph{Experimental results.} In \cref{fig:fze} we plot  2D histograms of the memorization and simplicity scores, $\hat{\mathcal{M}}_E$ and $\hat{\mathcal{S}}_E$. As it can be seen, the vast majority of the samples are not extractable at all, so we have $\hat{\mathcal{M}}_E = \hat{\mathcal{S}}_E = 0$. Some fraction of the training canaries are counterfactually extractable though, i.e., $\hat{\mathcal{M}}_E \gg 0$. 
To determine whether the technique proposed in \cref{sec:efficient} is actually identifying memorised samples, we now plot the Extractable Memorization and Simplicity scores of samples $E-G$ that were extractable only from the original model $f$, as well as the ``control'' samples $G$ that were extractable by the generalization baseline $f_G$. 
As expected, samples in $G$  have low  memorization scores $\hat{\mathcal{M}}_E$:
these answers can be extracted whether we train on them or not.
In contrast, samples in $E-G$ have memorization scores $\hat{\mathcal{M}}_E$ that vary between $0$ and $1$. Most of the samples are close to the line $\hat{\mathcal{S}}_E$ = $\hat{\mathcal{M}}_E$, indicating that the in-sample extractability is the only term contributing to $\hat{\mathcal{M}}_E$ (i.e., a model must see a sample at training time in order to extract it, and cannot extract it due to generalization only).

\input{figures/base_highres_pix2struct_fze_inpainted.tex}

%% file: figures/mem_vs_gen.tex
\begin{figure}[t!]
    \centering
    \begin{tikzpicture}[scale=0.65]
        \begin{axis}[
            width=13.5cm,
            height=8cm,
            ymode=log,
            ymin=1,
            xmin=0.5,
            xmax=11.5,
            ytick style={draw=none},
            xtick={1,2,3,4,5.5,6.5,7.5,8.5,10,11},
            xticklabels={$640\times640$,$960\times960$,$1280\times960$, $2560\times1920$,  $Large~0.6M$, $Large~1M$, $Base~0.6M$, $Base~1M$, $812\times812$,  $1064\times1064$ },
            x tick label style={rotate=35,anchor=east},
            point meta=rawy,
            legend columns=5, 
            legend cell align={left}, 
            legend image code/.code={\draw[#1, draw=none] (0cm,-0.1cm) rectangle (0.4cm,0.1cm);}            ]
            \addplot[ybar, bar width=0.35cm, bar shift=-0.35cm, draw=none,
                style={fill=orange!40}, nodes near coords, nodes near coords style={font=\footnotesize, xshift=-0.35cm}] table [x=x, y=e] {figures/rawdata/m_donut_results.txt}; 
            \addplot[ybar, bar width=0.35cm, bar shift=0cm, draw=none, nodes near coords, nodes near coords style={font=\footnotesize, xshift=0cm},
                style={fill=teal!40} ] table [x=x, y=g] {figures/rawdata/m_donut_results.txt}; 

            \addplot[ybar, bar width=0.35cm, bar shift=0.35cm, draw=none,
                style={fill=gray!40},nodes near coords, nodes near coords style={font=\footnotesize, xshift=0.35cm}] table [x=x, y=emg] {figures/rawdata/m_donut_results.txt}; 

            \addplot[ybar, bar width=0.35cm, bar shift=-0.35cm, draw=none,
                style={fill=orange!40}, nodes near coords, nodes near coords style={font=\footnotesize, xshift=-0.35cm}] table [x=x, y=e] {figures/rawdata/m_pix2struct_results.txt}; 
           \addplot[ybar, bar width=0.35cm, bar shift=0cm, draw=none,
                style={fill=teal!40}, nodes near coords, nodes near coords style={font=\footnotesize, xshift=0cm}] table [x=x, y=g] {figures/rawdata/m_pix2struct_results.txt}; 
            \addplot[ybar, bar width=0.35cm, bar shift=0.35cm, draw=none,
                style={fill=gray!40}, nodes near coords, nodes near coords style={font=\footnotesize, xshift=0.35cm}] table [x=x, y=emg] {figures/rawdata/m_pix2struct_results.txt}; 

            \addplot[ybar, bar width=0.35cm, bar shift=-0.35cm, draw=none,
                style={fill=orange!40}, nodes near coords, nodes near coords style={font=\footnotesize, xshift=-0.35cm}] table [x=x, y=e] {figures/rawdata/m_pali_results.txt}; 

           \addplot[ybar, bar width=0.35cm, bar shift=0cm, draw=none,
                style={fill=teal!40}, nodes near coords, nodes near coords style={font=\footnotesize, xshift=0cm}] table [x=x, y=g] {figures/rawdata/m_pali_results.txt}; 
                
            \addplot[ybar, bar width=0.35cm, bar shift=0.35cm, draw=none,
                style={fill=gray!40}, nodes near coords, nodes near coords style={font=\footnotesize, xshift=0.35cm}] table [x=x, y=emg] {figures/rawdata/m_pali_results.txt}; 

            \legend{$|E|$, $|G|$,$|E-G|$} 

        \end{axis}
    \end{tikzpicture}
    
    \begin{tikzpicture}[scale=0.5]
    \node (title) at (0,0) {\scriptsize{\hspace{-5cm}Donut}};
    \node (title) at (0,0) {\scriptsize{\hspace{1.5cm}Pix2Struct}};
    \node (title) at (0,0) {\scriptsize{\hspace{6.5cm}PaLI-3}};
    \end{tikzpicture}

    \caption{Extractability of answers for an attacker prompting the model with the original image from which the answer has been removed $I_i^{-a_i}$ and the original training question $Q_i$. The Y-axis is in logscale, therefore it overemphasizes the magnitued of lower values. PaLI-3 exhibits the lowest amount of extractable information in $M$.} \label{fig:table1}
    \end{figure}
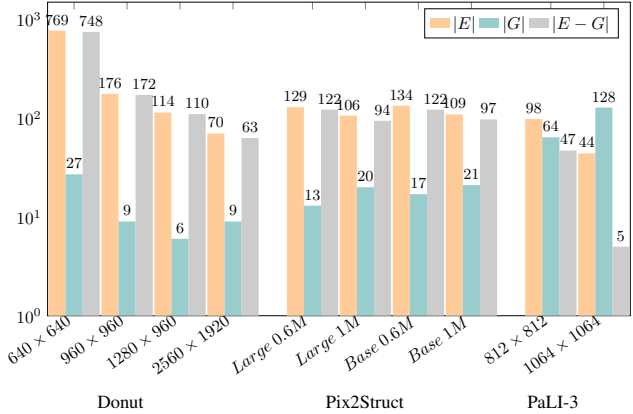

%% file: figures/mem_vs_gen_newviz.tex
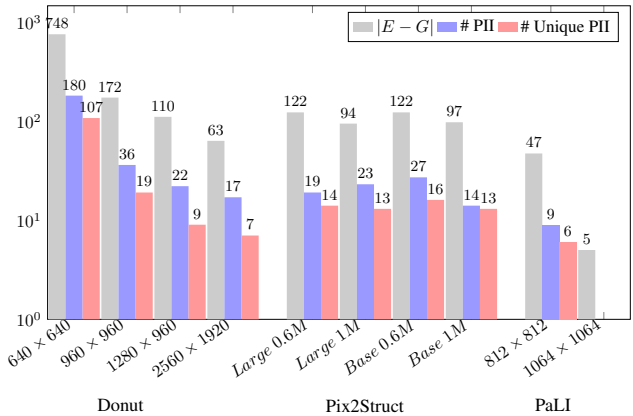
\begin{figure}[t!]
    \centering
    \begin{tikzpicture}[scale=0.65]
        \begin{axis}[
            width=13.5cm,
            height=8cm,
            ymode=log,
            ymin=1,
            xmin=0.5,
            xmax=11.5,
            ytick style={draw=none},
            xtick={1,2,3,4,5.5,6.5,7.5,8.5,10,11},
            xticklabels={$640\times640$,$960\times960$,$1280\times960$, $2560\times1920$,  $Large~0.6M$, $Large~1M$, $Base~0.6M$, $Base~1M$, $812\times812$,  $1064\times1064$ },
            x tick label style={rotate=35,anchor=east},
            point meta=rawy,
            legend columns=5, 
            legend cell align={left}, 
            legend image code/.code={\draw[#1, draw=none] (0cm,-0.1cm) rectangle (0.4cm,0.1cm);}            ]
            \addplot[ybar, bar width=0.35cm, bar shift=-0.35cm, draw=none,
                style={fill=gray!40}, nodes near coords, nodes near coords style={font=\footnotesize, xshift=-0.35cm}] table [x=x, y=emg] {figures/rawdata/m_donut_newviz.txt}; 
            \addplot[ybar, bar width=0.35cm, bar shift=0cm, draw=none, nodes near coords, nodes near coords style={font=\footnotesize, xshift=0cm},
                style={fill=blue!40} ] table [x=x, y=emg_pii] {figures/rawdata/m_donut_newviz.txt}; 

            \addplot[ybar, bar width=0.35cm, bar shift=0.35cm, draw=none,
                style={fill=red!40},nodes near coords, nodes near coords style={font=\footnotesize, xshift=0.35cm}] table [x=x, y=unique_emg_pii] {figures/rawdata/m_donut_newviz.txt}; 
            \addplot[ybar, bar width=0.35cm, bar shift=-0.35cm, draw=none,
                style={fill=gray!40}, nodes near coords, nodes near coords style={font=\footnotesize, xshift=-0.35cm}] table [x=x, y=emg] {figures/rawdata/m_pix2struct_newviz.txt}; 
           \addplot[ybar, bar width=0.35cm, bar shift=0cm, draw=none,
                style={fill=blue!40}, nodes near coords, nodes near coords style={font=\footnotesize, xshift=0cm}] table [x=x, y=emg_pii] {figures/rawdata/m_pix2struct_newviz.txt}; 
            \addplot[ybar, bar width=0.35cm, bar shift=0.35cm, draw=none,
                style={fill=red!40}, nodes near coords, nodes near coords style={font=\footnotesize, xshift=0.35cm}] table [x=x, y=unique_emg_pii] {figures/rawdata/m_pix2struct_newviz.txt}; 
            \addplot[ybar, bar width=0.35cm, bar shift=-0.35cm, draw=none,
                style={fill=gray!40}, nodes near coords, nodes near coords style={font=\footnotesize, xshift=-0.35cm}] table [x=x, y=emg] {figures/rawdata/m_pali_newviz.txt}; 
           \addplot[ybar, bar width=0.35cm, bar shift=0cm, draw=none,
                style={fill=blue!40}, nodes near coords, nodes near coords style={font=\footnotesize, xshift=0cm}] table [x=x, y=emg_pii] {figures/rawdata/m_pali_newviz.txt}; 
                
            \addplot[ybar, bar width=0.35cm, bar shift=0.35cm, draw=none,
                style={fill=red!40}, nodes near coords, nodes near coords style={font=\footnotesize, xshift=0.35cm}] table [x=x, y=unique_emg_pii] {figures/rawdata/m_pali_newviz.txt};

            \legend{$|E-G|$, \# PII, \# Unique PII} 

        \end{axis}
    \end{tikzpicture}
    
    \begin{tikzpicture}[scale=0.5]
    \node (title) at (0,0) {\scriptsize{\hspace{-5cm}Donut}};
    \node (title) at (0,0) {\scriptsize{\hspace{1.5cm}Pix2Struct}};
    \node (title) at (0,0) {\scriptsize{\hspace{6.5cm}PaLI}};
    \end{tikzpicture}
    
    \caption{Amount of samples in $M$ that are PII, and amount of samples that are unique PIIs when querying the model with $(I^{-a}, Q)$. \label{fig:table2}}
    \end{figure}

%% file: figures/base_highres_pix2struct_fze_inpainted.tex
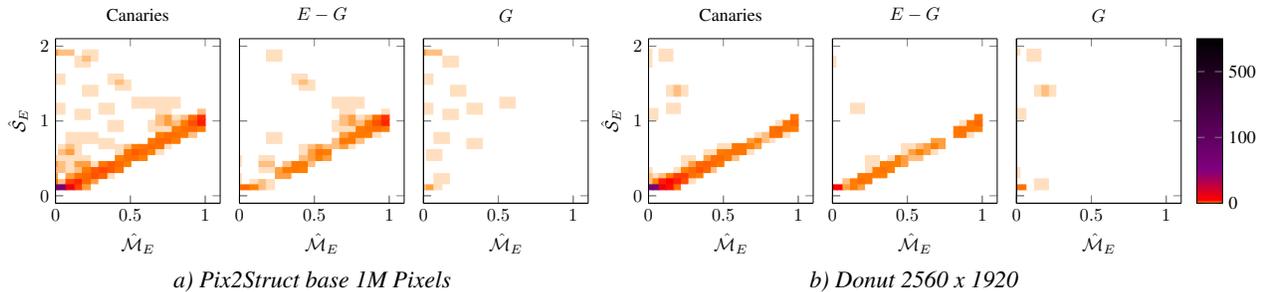
\begin{figure*}[t]
    \centering
    \begin{tikzpicture}[scale=0.68]
        \begin{axis}[
        title=Canaries,
        width=4.8cm,
        height=4.8cm,
        xmin=0,
        xmax=1.1,
        ymin=-0.1,
        ymax=2.1,
        ylabel=$\hat{\mathcal{S}}_E$,
        xlabel=$\hat{\mathcal{M}}_E$,
        view={0}{90},
        point meta max=500,
        colormap={new}{color(0)=(white);color(1)=(orange);color(10)=(red);color(100)=(violet);color(500)=(black)}
        ]
            \addplot3[surf, shader=flat ] table[x=x,y=y,z=fz] {figures/base_highres_pix2struct_fze_inpainted.txt};
        \end{axis}
    \end{tikzpicture}
    \begin{tikzpicture}[scale=0.68]
        \begin{axis}[
        title= $E - G$,
        width=4.8cm,
        height=4.8cm,
        xmin=0,
        xmax=1.1,
        ymin=-0.1,
        ymax=2.1,
        yticklabels=none, 
        xlabel=$\hat{\mathcal{M}}_E$,
        view={0}{90},
        point meta max=500,
        colormap={new}{color(0)=(white);color(1)=(orange);color(10)=(red);color(100)=(violet);color(500)=(black)}
        ]
            \addplot3[surf, shader=flat] table[x=x,y=y,z=e] {figures/base_highres_pix2struct_fze_inpainted.txt};
        \end{axis}
    \end{tikzpicture}
    \begin{tikzpicture}[scale=0.68]
        \begin{axis}[
        title= $G$,
        width=4.8cm,
        height=4.8cm,
        xmin=0,
        xmax=1.1,
        ymin=-0.1,
        ymax=2.1,
        yticklabels=none, 
        xlabel=$\hat{\mathcal{M}}_E$,
        view={0}{90},
        point meta max=500,
        colormap={new}{color(0)=(white);color(1)=(orange);color(10)=(red);color(100)=(violet);color(500)=(black)}
        ]
            \addplot3[surf, shader=flat] table[x=x,y=y,z=g] {figures/base_highres_pix2struct_fze_inpainted.txt};
        \end{axis}
        
    \end{tikzpicture}
    \begin{tikzpicture}[scale=0.68]
        \begin{axis}[
        title= Canaries,
        width=4.8cm,
        height=4.8cm,
        xmin=0,
        xmax=1.1,
        ymin=-0.1,
        ymax=2.1,
        ylabel=$\hat{\mathcal{S}}_E$,
        xlabel=$\hat{\mathcal{M}}_E$,
        view={0}{90},
        point meta max=500,
        colormap={new}{color(0)=(white);color(1)=(orange);color(10)=(red);color(100)=(violet);color(500)=(black)}
        ]
            \addplot3[surf, shader=flat ] table[x=x,y=y,z=fz] {figures/res0_donut_fze_inpainted.txt};
        \end{axis}
    \end{tikzpicture}
    \begin{tikzpicture}[scale=0.68]
        \begin{axis}[
        title= $E - G$,
        width=4.8cm,
        height=4.8cm,
        xmin=0,
        xmax=1.1,
        ymin=-0.1,
        ymax=2.1,
        yticklabels=none, 
        xlabel=$\hat{\mathcal{M}}_E$,
        view={0}{90},
        point meta max=500,
        colormap={new}{color(0)=(white);color(1)=(orange);color(10)=(red);color(100)=(violet);color(500)=(black)}
        ]
            \addplot3[surf, shader=flat ] table[x=x,y=y,z=e] {figures/res0_donut_fze_inpainted.txt};
        \end{axis}
    
    \end{tikzpicture}
    \begin{tikzpicture}[scale=0.68]
        \begin{axis}[
        title= $G$,
        width=4.8cm,
        height=4.8cm,
        xmin=0,
        xmax=1.1,
        ymin=-0.1,
        ymax=2.1,
        yticklabels=none,
        xlabel=$\hat{\mathcal{M}}_E$,
        view={0}{90},
        colorbar,
        point meta max=500,
        colorbar style={yticklabels={0,0,100,500}},
        colormap={new}{color(0)=(white);color(1)=(orange);color(10)=(red);color(100)=(violet);color(500)=(black)}
        ]
            \addplot3[surf, shader=flat] table[x=x,y=y,z=g] {figures/res0_donut_fze_inpainted.txt};
        \end{axis}
    \end{tikzpicture}
    
    \begin{tikzpicture}[scale=0.5]
    \node (title) at (0,-0) {\footnotesize{\hspace{-8.5cm}\textit{a) Pix2Struct base 1M Pixels}}};
    \node (title) at (0,-0) {\footnotesize{\hspace{7.5cm}\textit{b) Donut 2560 x 1920}}};
    \end{tikzpicture}
    
    \caption{Distributions of the $\hat{\mathcal{M}}_E$ and $\hat{\mathcal{S}}_E$ scores for all the canaries, $E-G$ and $G$ for both Pix2Struct base 1M Pixels (three panels on the left) and Donut 2560 x 1920 (three panels on the right). Samples in $E-G$ have high memorization scores, while samples in $G$ do not.} \label{fig:fze} 
\end{figure*}

%% file: 4_reliance_on_context.tex
\section{Ablations on the Extraction Context} \label{sec:ctx_importance}

So far, we studied the extractability of an answer $a$ assuming knowledge of all other parts of an input. We now relax this assumption to both gain further insights into the factors influencing extractability, and, in some cases, to simulate more realistic attack scenarios in which perfect knowledge of the context $(I^{-a},Q)$ is not available. Indeed, while perfect knowledge of the context is unlikely in many cases, it is possible for an attacker to craft an approximation of the context (e.g., because the information they are seeking is contained in documents with a known or fixed structure, like driving licences or forms available online).

Before delving in the results, we point out that just like modifying the way a LLM is prompted can modify its output significantly, changing the way the VLMs are prompted changes which samples are extractable. For this reason, in few cases, the amount of extractable samples may increase with respect to the baseline scenario we considered so far, especially for cases in which the generalization baseline is weakened by the reduced information contained in the approximation of the context. 
\subsection{No Text in the Image} \label{sec:surrounding-text}
For LLMs, prior work has shown that prompting a model with the prefix of a memorized string is a reliable way of extracting data~\cite{carlini2023quantifying,tirumala2022memorization}. Yet, for Document-Based VQA systems it is unclear whether the models actually needs to read any surrounding text in a document in order to recall the answer.
For this reason, we study the case in which \emph{all} text is removed from the image $I$.  If the model can still respond correctly, it indicates the model is relying on the question and non-textual features (e.g., layout, presence of icons or images etc.) in order to regurgitate the answer.  
This experiment also represents a practical threat model where the attacker knows the layout of a document (e.g., because it is a form available online or a document with a fixed structure like driving licences or ID cards) but has little to no knowledge about its contents.

\textit{Results}: Figure \ref{label:experiments} shows that in case of Donut and Pix2Struct, the absence of text in the image significantly reduces the ability of the model to return the correct answer. In case of Donut the amount of samples in $M$ is 26. Pix2Struct shows a similar decrease from about 94 to 27. The amount of PIIs returned is also significantly reduced, and consisting mostly of highly repeated PIIs (more than $6$ times). In the case of PaLI-3, we also observe the model responds correctly to answers requiring general knowledge (e.g., the name of chemical substances from their symbols contained in the questions, names of animal species portrayed in pictures contained in the document). The increase in the amount of extractable answers may be related to the fact that, when the extraction fails, a typical pattern is for the model to read another part of the document. When no text is present, it is easier for the model to retrieve the information from the general knowledge it acquired at pre-training time.
For PaLI-3, no PII is extracted.

\begin{boxK}
\textbf{Reliance on surrounding text}: The lack of any text in the document significantly reduces the ability to extract unique PIIs.
\end{boxK}

\subsection{Imperfect Knowledge of the Training Question} 

To understand whether the model is 
memorizing an association between the exact question $Q$ and answer $a$, we measure whether we can extract the answer when the question is paraphrased. We create paraphrases $Q'$ of $Q$ and extract the answers using $(I^{-a},Q')$. To this end, we use PaLM2 \cite{anil2023palm} to create a paraphrased question for each canary question. An example of paraphrase is the following: if the question $Q$ is``What is the address shown in the document?'', then the paraphrase $Q'$ can be ``What is the street name and city shown in the document?''. This experiment also reflects the setting in which the attacker does not know the exact phrasing of the training question $Q$ and approximates it with their own words. 

\textit{Results}: Figure \ref{label:experiments} shows that the number of extracted answers significantly drops, but is still non-negligible. For both Pix2Struct and Donut we observe several unique PIIs are extractable (e.g., names of individuals, serial numbers of tickets and travel destinations).
The extractability increases in the case of PaLI-3, but is again related to questions probing general knowledge and reveal no PII.

\begin{boxK}
\textbf{Robustness to paraphrasing of $Q$}: 
 Uncertainty about the exact phrasing of a question that queries PII does not prevent extraction of sensitive information, but can reduce the amount of extractable samples.
\end{boxK}

\subsection{Robustness to Image Perturbations} \label{sec:perturbed-image}
An attacker may be able to craft a document similar to the one originally used for training, but the scanning procedure naturally induces some small visual differences that may influence the extractability of the answers (e.g. brightness changes, small rotations or translations).  For this reason we consider the case in which the original context $I^{-a}$ is perturbed with augmentations that reflect plausible differences that may incur between the training and adversarially crafted document scans. For this purpose, we consider the following augmentations: 1) brightness change: we increase ($\times 1.3, \times 2$) or decrease ($\times 0.8,\times 0.5$) the brightness of the document; 2) small rotations: we randomly rotate by $\pm 5$ or $\pm 10$ degrees; 3) small translations: we randomly shift the image by $\pm 20$ and $\pm 100$ pixels along both axes.

\emph{Results:} In Figure \ref{label:experiments}, we can see that brightness changes can indeed reduce the amount of extractable information, but the amount of extractable samples is still significantly high. In most cases, the stronger the change in brightness, the less the answer is extractable. However, a substantial amount of samples remains extractable, especially with respect to the context perturbations considered in the previous sections. Rotating or translating the image has a stronger adverse effect on the extractability of answers, indicating that spatial information plays a more important role for extractability than the intensity information. Notice, the amount of extractable samples under image perturbations is significantly larger than the amount extractable when the question is paraphrased, indicating that precise knowledge of the question $Q$ is more important for an extraction attack than precise knowledge of the original scan $I^{-a}$.
This also suggests that extractability is more likely to be triggered in the presence of the training question than in presence of the input image $I^{-a}$. 

\begin{boxK}
    \textbf{Robustness to Image Perturbations} The amount of extractable samples is relatively robust to brightness perturbations and less to spatial transformations. An adversary does not need to reproduce a perfect copy of the original training image to extract the answer.   
\end{boxK}

\subsection{Permuting Modalities} \label{sec:shuffling}
Document-Based VQA systems contain both a visual component and a language component, each of which are fine-tuned on the training data. Extensive evidence has been provided that each of these components can memorise training data \emph{in isolation} \cite{feldman2019memorization, lukasik2023larger, carlini2022membership, carlini2019secret}. Therefore an interesting question is whether it is possible for a multimodal model to extract the answers independently of one of the two input modalities. For this reason, we consider two experiments that randomise the relationship between the two input modalities.

\textbf{Extractability based on questions only.} 
At inference time, we feed the model a partial image with an unrelated question $(I_j^{-a_j}, Q_i)$, where $i \neq j$ and there is no training sample with question $Q_i$ applied to image $I_j$, and the correct answer to question $Q_i$ does not appear in the text of image $I_j$.
This experiment evaluates the ability of the model to respond solely based on the question and reflects the case in which the attacker does not know the image $I_i$ at all.\footnote{We have also tried replacing the input image with constant intensity value set to black, white or the average value of $I_i$. No answer was extractable in this case, perhaps because such images are too far out-of-distribution.}

\input{figures/results.tex}

\textit{Results}: 
In the setting where we try to extract the original answer $a_i$, as visible in the Shuffling column in \cref{label:experiments}, we can extract only 4 answers in case of Donut, and  21 in case of Pix2Struct. Among all the samples in $M$, we can also find some sensitive samples containing area codes, names of individuals and dates in which the documents were issued. The sensitive samples are also repeated only once or at most twice in the model's training set. While 2 answers can be extracted for PaLI-3, no PII was extracted.

\textbf{Extractability based on images only.}  As in the previous experiment, we provide the model with a partial input image and an unrelated question that does not contain an answer within the image. We then measure whether we can extract an answer to one of the questions that was asked about this image during training.
We find no extractable answers in this setting, which suggests that the question plays a more predominant role in the extraction.

\begin{boxK}
    \textbf{Dependency of extractability on modalities} In few cases, the model can leverage the language component alone to extract sensitive answers. If the training answers are not present in the image modality and the question was not seen at training time for a specific document, the image alone is not sufficient to extract any memorized answer.  
\end{boxK}

%% file: figures/results.tex
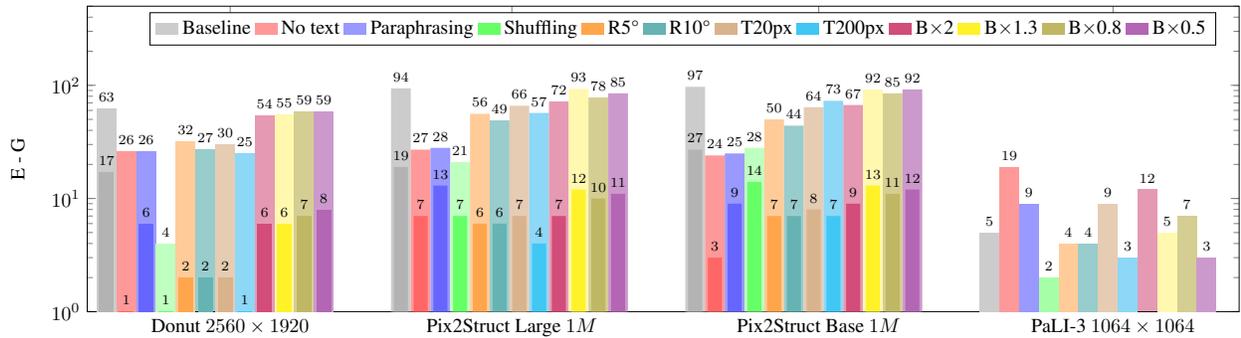
\begin{figure*}[t]
    \centering
    \begin{tikzpicture}[scale=0.75]
        \begin{axis}[
            width=22cm,
            height=7cm,
            ymode=log,
            ylabel=E - G,
            ymin=1,
            ymax=500,
            xmin=0.3,
            xmax=14
            ,
            xtick={2, 5.35, 9, 12.5}, 
            xticklabels={Donut $2560 \times 1920$, Pix2Struct Large $1M$, Pix2Struct Base $1M$, PaLI-3 $1064 \times 1064$}, 
            point meta=rawy,
            legend columns=12, 
            legend style={xshift=0cm},
            legend image code/.code={\draw[#1, draw=none] (0cm,-0.1cm) rectangle (0.4cm,0.1cm);}
            ]
            \addplot[ybar, bar width=0.35cm, bar shift=-0.7cm, draw=none,
                style={fill=gray!40}, nodes near coords, nodes near coords style={font=\scriptsize, xshift=-0.7cm}] table [x=x, y=baseline] {figures/donut_results.txt}; 
            \addplot[ybar, bar width=0.35cm, bar shift=-0.35cm, draw=none,
                style={fill=red!40}, nodes near coords, nodes near coords style={font=\scriptsize, xshift=-0.35cm}] table [x=x, y=notext] {figures/donut_results.txt};
            \addplot[ybar, bar width=0.35cm, draw=none, 
                style={fill=blue!40}, nodes near coords, nodes near coords style={font=\scriptsize, xshift=0cm}] table [x=x, y=paraphrases] {figures/donut_results.txt};
            \addplot[ybar, bar width=0.35cm, bar shift=0.35cm, draw=none,
                style={fill=green!25}, nodes near coords, nodes near coords style={font=\scriptsize, xshift=0.35cm}] table [x=x, y=shuffle] {figures/donut_results.txt};

            \addplot[ybar, bar width=0.25cm, bar shift=-0.7cm, draw=none,
                style={fill=gray!60}, nodes near coords, nodes near coords style={font=\scriptsize, xshift=-0.7cm}] table [x=x, y=baseline_pii] {figures/donut_results.txt}; 
            \addplot[ybar, bar width=0.25cm, bar shift=-0.35cm, draw=none,
                style={fill=red!60}, nodes near coords, nodes near coords style={font=\scriptsize, xshift=-0.35cm}] table [x=x, y=notext_pii] {figures/donut_results.txt};
            \addplot[ybar, bar width=0.25cm, draw=none, 
                style={fill=blue!60}, nodes near coords, nodes near coords style={font=\scriptsize, xshift=0cm}] table [x=x, y=paraphrases_pii] {figures/donut_results.txt};    
            \addplot[ybar, bar width=0.25cm, bar shift=0.35cm, draw=none,
                style={fill=green!60}, nodes near coords, nodes near coords style={font=\scriptsize, xshift=0.35cm}] table [x=x, y=shuffle_pii] {figures/donut_results.txt};
            \addplot[ybar, bar width=0.35cm, bar shift=0.7cm, draw=none,
                style={fill=orange!40}, nodes near coords, nodes near coords style={font=\scriptsize, xshift=0.7cm}] table [x=x, y=rot5] {figures/i_perturb/i_donut_results.txt}; 
            \addplot[ybar, bar width=0.25cm, bar shift=0.7cm, draw=none,
                style={fill=orange!70}, nodes near coords, nodes near coords style={font=\scriptsize, xshift=0.7cm}] table [x=x, y=rot5_pii] {figures/i_perturb/i_donut_results.txt}; 
            \addplot[ybar, bar width=0.35cm, bar shift=1.05cm, draw=none, nodes near coords, nodes near coords style={font=\scriptsize, xshift=1.05cm},
                style={fill=teal!40} ] table [x=x, y=rot10] {figures/i_perturb/i_donut_results.txt}; 
            \addplot[ybar, bar width=0.25cm, bar shift=1.05cm, draw=none, nodes near coords, nodes near coords style={font=\scriptsize, xshift=1.05cm},
                style={fill=teal!60} ] table [x=x, y=rot10_pii] {figures/i_perturb/i_donut_results.txt}; 
            \addplot[ybar, bar width=0.35cm, bar shift=1.40cm, draw=none,
                style={fill=brown!40},nodes near coords, nodes near coords style={font=\scriptsize, xshift=1.40cm}] table [x=x, y=t20] {figures/i_perturb/i_donut_results.txt}; 
            \addplot[ybar, bar width=0.25cm, bar shift=1.40cm, draw=none,
                style={fill=brown!60},nodes near coords, nodes near coords style={font=\scriptsize, xshift=1.40cm}] table [x=x, y=t20_pii] {figures/i_perturb/i_donut_results.txt}; 
            \addplot[ybar, bar width=0.35cm, bar shift=1.75cm, draw=none,
                style={fill=cyan!40}, nodes near coords, nodes near coords style={font=\scriptsize, xshift=1.75cm}] table [x=x, y=t100] {figures/i_perturb/i_donut_results.txt}; 
            \addplot[ybar, bar width=0.25cm, bar shift=1.75cm, draw=none,
                style={fill=cyan!60}, nodes near coords, nodes near coords style={font=\scriptsize, xshift=1.75cm}] table [x=x, y=t100_pii] {figures/i_perturb/i_donut_results.txt};

            \addplot[ybar, bar width=0.35cm, bar shift=2.1cm, draw=none,
                style={fill=purple!40}, nodes near coords, nodes near coords style={font=\scriptsize, xshift=2.1cm}] table [x=x, y=b2] {figures/i_perturb/i_donut_results.txt}; 

            \addplot[ybar, bar width=0.25cm, bar shift=2.1cm, draw=none,
                style={fill=purple!70}, nodes near coords, nodes near coords style={font=\scriptsize, xshift=2.1cm}] table [x=x, y=b2_pii] {figures/i_perturb/i_donut_results.txt}; 

            \addplot[ybar, bar width=0.35cm, bar shift=2.45cm, draw=none, nodes near coords, nodes near coords style={font=\scriptsize, xshift=2.45cm},
                style={fill=yellow!40} ] table [x=x, y=b13] {figures/i_perturb/i_donut_results.txt}; 
            \addplot[ybar, bar width=0.25cm, bar shift=2.45cm, draw=none, nodes near coords, nodes near coords style={font=\scriptsize, xshift=2.45cm},
                style={fill=yellow!90} ] table [x=x, y=b13_pii] {figures/i_perturb/i_donut_results.txt}; 

            \addplot[ybar, bar width=0.35cm, bar shift=2.8cm, draw=none,
                style={fill=olive!40},nodes near coords, nodes near coords style={font=\scriptsize, xshift=2.8cm}] table [x=x, y=b08] {figures/i_perturb/i_donut_results.txt}; 
            \addplot[ybar, bar width=0.25cm, bar shift=2.8cm, draw=none,
                style={fill=olive!60},nodes near coords, nodes near coords style={font=\scriptsize, xshift=2.8cm}] table [x=x, y=b08_pii] {figures/i_perturb/i_donut_results.txt}; 

            \addplot[ybar, bar width=0.35cm, bar shift=3.15cm, draw=none,
                style={fill=violet!40}, nodes near coords, nodes near coords style={font=\scriptsize, xshift=3.15cm}] table [x=x, y=b05] {figures/i_perturb/i_donut_results.txt}; 
            \addplot[ybar, bar width=0.25cm, bar shift=3.15cm, draw=none,
                style={fill=violet!60}, nodes near coords, nodes near coords style={font=\scriptsize, xshift=3.15cm}] table [x=x, y=b05_pii] {figures/i_perturb/i_donut_results.txt};

            \addplot[ybar, bar width=0.35cm, bar shift=-0.7cm, draw=none,
                style={fill=gray!40}, nodes near coords, nodes near coords style={font=\scriptsize, xshift=-0.7cm}] table [x=x, y=baseline] {figures/pix2struct_results.txt};
            \addplot[ybar, bar width=0.25cm, bar shift=-0.7cm, draw=none,
                style={fill=gray!60}, nodes near coords, nodes near coords style={font=\scriptsize, xshift=-0.7cm}] table [x=x, y=baseline_pii] {figures/pix2struct_results.txt};
            \addplot[ybar, bar width=0.35cm, bar shift=-0.35cm, draw=none,
                style={fill=red!40}, nodes near coords, nodes near coords style={font=\scriptsize, xshift=-0.35cm}] table [x=x, y=notext] {figures/pix2struct_results.txt};
             \addplot[ybar, bar width=0.25cm, bar shift=-0.35cm, draw=none,
                style={fill=red!60}, nodes near coords, nodes near coords style={font=\scriptsize, xshift=-0.35cm}] table [x=x, y=notext_pii] {figures/pix2struct_results.txt};
            \addplot[ybar, bar width=0.35cm, draw=none, 
                style={fill=blue!40}, nodes near coords, nodes near coords style={font=\scriptsize, xshift=0cm}] table [x=x, y=paraphrases] {figures/pix2struct_results.txt};
            \addplot[ybar, bar width=0.25cm, draw=none, 
                style={fill=blue!60}, nodes near coords, nodes near coords style={font=\scriptsize, xshift=0cm}] table [x=x, y=paraphrases_pii] {figures/pix2struct_results.txt};
            \addplot[ybar, bar width=0.35cm, bar shift=0.35cm, draw=none,
                style={fill=green!25}, nodes near coords, nodes near coords style={font=\scriptsize, xshift=0.35cm}] table [x=x, y=shuffle] {figures/pix2struct_results.txt};
            \addplot[ybar, bar width=0.25cm, bar shift=0.35cm, draw=none,
                style={fill=green!60}, nodes near coords, nodes near coords style={font=\scriptsize, xshift=0.35cm}] table [x=x, y=shuffle_pii] {figures/pix2struct_results.txt};

            \addplot[ybar, bar width=0.35cm, bar shift=0.7cm, draw=none,
                style={fill=orange!40}, nodes near coords, nodes near coords style={font=\scriptsize, xshift=0.7cm}] table [x=x, y=rot5] {figures/i_perturb/i_pix2struct_results.txt}; 
            \addplot[ybar, bar width=0.25cm, bar shift=0.7cm, draw=none,
                style={fill=orange!70}, nodes near coords, nodes near coords style={font=\scriptsize, xshift=0.7cm}] table [x=x, y=rot5_pii] {figures/i_perturb/i_pix2struct_results.txt}; 
            \addplot[ybar, bar width=0.35cm, bar shift=1.05cm, draw=none, nodes near coords, nodes near coords style={font=\scriptsize, xshift=1.05cm},
                style={fill=teal!40} ] table [x=x, y=rot10] {figures/i_perturb/i_pix2struct_results.txt}; 
            \addplot[ybar, bar width=0.25cm, bar shift=1.05cm, draw=none, nodes near coords, nodes near coords style={font=\scriptsize, xshift=1.05cm},
                style={fill=teal!60} ] table [x=x, y=rot10_pii] {figures/i_perturb/i_pix2struct_results.txt}; 
            \addplot[ybar, bar width=0.35cm, bar shift=1.40cm, draw=none,
                style={fill=brown!40},nodes near coords, nodes near coords style={font=\scriptsize, xshift=1.40cm}] table [x=x, y=t20] {figures/i_perturb/i_pix2struct_results.txt}; 
            \addplot[ybar, bar width=0.25cm, bar shift=1.40cm, draw=none,
                style={fill=brown!60},nodes near coords, nodes near coords style={font=\scriptsize, xshift=1.40cm}] table [x=x, y=t20_pii] {figures/i_perturb/i_pix2struct_results.txt}; 
            \addplot[ybar, bar width=0.35cm, bar shift=1.75cm, draw=none,
                style={fill=cyan!40}, nodes near coords, nodes near coords style={font=\scriptsize, xshift=1.75cm}] table [x=x, y=t100] {figures/i_perturb/i_pix2struct_results.txt}; 
            \addplot[ybar, bar width=0.25cm, bar shift=1.75cm, draw=none,
                style={fill=cyan!60}, nodes near coords, nodes near coords style={font=\scriptsize, xshift=1.75cm}] table [x=x, y=t100_pii] {figures/i_perturb/i_pix2struct_results.txt};

            \addplot[ybar, bar width=0.35cm, bar shift=2.1cm, draw=none,
                style={fill=purple!40}, nodes near coords, nodes near coords style={font=\scriptsize, xshift=2.1cm}] table [x=x, y=b2] {figures/i_perturb/i_pix2struct_results.txt}; 

            \addplot[ybar, bar width=0.25cm, bar shift=2.1cm, draw=none,
                style={fill=purple!70}, nodes near coords, nodes near coords style={font=\scriptsize, xshift=2.1cm}] table [x=x, y=b2_pii] {figures/i_perturb/i_pix2struct_results.txt}; 

            \addplot[ybar, bar width=0.35cm, bar shift=2.45cm, draw=none, nodes near coords, nodes near coords style={font=\scriptsize, xshift=2.45cm},
                style={fill=yellow!40} ] table [x=x, y=b13] {figures/i_perturb/i_pix2struct_results.txt}; 
            \addplot[ybar, bar width=0.25cm, bar shift=2.45cm, draw=none, nodes near coords, nodes near coords style={font=\scriptsize, xshift=2.45cm},
                style={fill=yellow!90} ] table [x=x, y=b13_pii] {figures/i_perturb/i_pix2struct_results.txt}; 

            \addplot[ybar, bar width=0.35cm, bar shift=2.8cm, draw=none,
                style={fill=olive!40},nodes near coords, nodes near coords style={font=\scriptsize, xshift=2.8cm}] table [x=x, y=b08] {figures/i_perturb/i_pix2struct_results.txt}; 
            \addplot[ybar, bar width=0.25cm, bar shift=2.8cm, draw=none,
                style={fill=olive!60},nodes near coords, nodes near coords style={font=\scriptsize, xshift=2.8cm}] table [x=x, y=b08_pii] {figures/i_perturb/i_pix2struct_results.txt}; 

            \addplot[ybar, bar width=0.35cm, bar shift=3.15cm, draw=none,
                style={fill=violet!40}, nodes near coords, nodes near coords style={font=\scriptsize, xshift=3.15cm}] table [x=x, y=b05] {figures/i_perturb/i_pix2struct_results.txt}; 
            \addplot[ybar, bar width=0.25cm, bar shift=3.15cm, draw=none,
                style={fill=violet!60}, nodes near coords, nodes near coords style={font=\scriptsize, xshift=3.15cm}] table [x=x, y=b05_pii] {figures/i_perturb/i_pix2struct_results.txt};


            \addplot[ybar, bar width=0.35cm, bar shift=-0.7cm, draw=none,
                style={fill=gray!40}, nodes near coords, nodes near coords style={font=\scriptsize, xshift=-0.7cm}] table [x=x, y=baseline] {figures/pali_results.txt};
            \addplot[ybar, bar width=0.35cm, bar shift=-0.35cm, draw=none,
                style={fill=red!40}, nodes near coords, nodes near coords style={font=\scriptsize, xshift=-0.35cm}] table [x=x, y=notext] {figures/pali_results.txt};
            \addplot[ybar, bar width=0.35cm, draw=none, 
                style={fill=blue!40}, nodes near coords, nodes near coords style={font=\scriptsize, xshift=0cm}] table [x=x, y=paraphrases] {figures/pali_results.txt};
            \addplot[ybar, bar width=0.35cm, bar shift=0.35cm, draw=none,
                style={fill=green!35}, nodes near coords, nodes near coords style={font=\scriptsize, xshift=0.35cm}] table [x=x, y=shuffle] {figures/pali_results.txt};

            \addplot[ybar, bar width=0.35cm, bar shift=0.7cm, draw=none,
                style={fill=orange!40}, nodes near coords, nodes near coords style={font=\scriptsize, xshift=0.7cm}] table [x=x, y=rot5] {figures/i_perturb/i_pali_results.txt}; 
            \addplot[ybar, bar width=0.35cm, bar shift=0.7cm, draw=none,
                style={fill=orange!70}, nodes near coords, nodes near coords style={font=\scriptsize, xshift=0.7cm}] table [x=x, y=rot5_pii] {figures/i_perturb/i_pali_results.txt}; 
            \addplot[ybar, bar width=0.35cm, bar shift=1.05cm, draw=none, nodes near coords, nodes near coords style={font=\scriptsize, xshift=1.05cm},
                style={fill=teal!40} ] table [x=x, y=rot10] {figures/i_perturb/i_pali_results.txt}; 
            \addplot[ybar, bar width=0.35cm, bar shift=1.05cm, draw=none, nodes near coords, nodes near coords style={font=\scriptsize, xshift=1.05cm},
                style={fill=teal!60} ] table [x=x, y=rot10_pii] {figures/i_perturb/i_pali_results.txt}; 
            \addplot[ybar, bar width=0.35cm, bar shift=1.40cm, draw=none,
                style={fill=brown!40},nodes near coords, nodes near coords style={font=\scriptsize, xshift=1.40cm}] table [x=x, y=t20] {figures/i_perturb/i_pali_results.txt}; 
            \addplot[ybar, bar width=0.35cm, bar shift=1.40cm, draw=none,
                style={fill=brown!60},nodes near coords, nodes near coords style={font=\scriptsize, xshift=1.40cm}] table [x=x, y=t20_pii] {figures/i_perturb/i_pali_results.txt}; 
            \addplot[ybar, bar width=0.35cm, bar shift=1.75cm, draw=none,
                style={fill=cyan!40}, nodes near coords, nodes near coords style={font=\scriptsize, xshift=1.75cm}] table [x=x, y=t100] {figures/i_perturb/i_pali_results.txt}; 
            \addplot[ybar, bar width=0.35cm, bar shift=1.75cm, draw=none,
                style={fill=cyan!60}, nodes near coords, nodes near coords style={font=\scriptsize, xshift=1.75cm}] table [x=x, y=t100_pii] {figures/i_perturb/i_pali_results.txt};

            \addplot[ybar, bar width=0.35cm, bar shift=2.1cm, draw=none,
                style={fill=purple!40}, nodes near coords, nodes near coords style={font=\scriptsize, xshift=2.1cm}] table [x=x, y=b2] {figures/i_perturb/i_pali_results.txt}; 

            \addplot[ybar, bar width=0.35cm, bar shift=2.1cm, draw=none,
                style={fill=purple!70}, nodes near coords, nodes near coords style={font=\scriptsize, xshift=2.1cm}] table [x=x, y=b2_pii] {figures/i_perturb/i_pali_results.txt}; 

            \addplot[ybar, bar width=0.35cm, bar shift=2.45cm, draw=none, nodes near coords, nodes near coords style={font=\scriptsize, xshift=2.45cm},
                style={fill=yellow!40} ] table [x=x, y=b13] {figures/i_perturb/i_pali_results.txt}; 
            \addplot[ybar, bar width=0.35cm, bar shift=2.45cm, draw=none, nodes near coords, nodes near coords style={font=\scriptsize, xshift=2.45cm},
                style={fill=yellow!90} ] table [x=x, y=b13_pii] {figures/i_perturb/i_pali_results.txt}; 

            \addplot[ybar, bar width=0.35cm, bar shift=2.8cm, draw=none,
                style={fill=olive!40},nodes near coords, nodes near coords style={font=\scriptsize, xshift=2.8cm}] table [x=x, y=b08] {figures/i_perturb/i_pali_results.txt}; 
            \addplot[ybar, bar width=0.35cm, bar shift=2.8cm, draw=none,
                style={fill=olive!60},nodes near coords, nodes near coords style={font=\scriptsize, xshift=2.8cm}] table [x=x, y=b08_pii] {figures/i_perturb/i_pali_results.txt}; 

            \addplot[ybar, bar width=0.35cm, bar shift=3.15cm, draw=none,
                style={fill=violet!40}, nodes near coords, nodes near coords style={font=\scriptsize, xshift=3.15cm}] table [x=x, y=b05] {figures/i_perturb/i_pali_results.txt}; 
            \addplot[ybar, bar width=0.35cm, bar shift=3.15cm, draw=none,
                style={fill=violet!60}, nodes near coords, nodes near coords style={font=\scriptsize, xshift=3.15cm}] table [x=x, y=b05_pii] {figures/i_perturb/i_pali_results.txt};

            \legend{Baseline, No text, Paraphrasing  ,,,,,Shuffling,, R5°,, R10°,, T20px,, T200px,, B$\times$2,, B$\times$1.3,,B$\times$0.8,,B$\times$0.5}
        \end{axis}
    \end{tikzpicture}

    \caption{Extractability of answers when the context does not contain the text (No Text), the question is paraphrased (Paraphrasing),  or not related to the image but the model still responds correctly (Shuffling), the image undergoes rotations (R5* and R10*), translations (T20px, T100px) and when brightness is changed by a mutliplicative factor (B$\times$2, 1.3, 0.8 or 0.5).   Darker colors indicate the number of PII samples that are extractable. Y-axis is in logscale. Across all deployable models, PaLI-3 exhibits the lowest amount of extractable information. \label{label:experiments}}
\end{figure*}

%% file: 5_mitigation.tex
\section{Defenses} \label{sec:countermeasures}
To conclude our study, we consider various mitigation strategies and measure their impact on memorization and generalization capabilities of the models (by computing the ANLS \cite{DocVQA_2021_WACV} on a held-out test set):
\begin{compactitem}
    \item \textbf{Inference Time Paraphrasing (ITP)}, similar to \cite{somepalli2023diffusion} we consider its effectiveness as a defense strategy. 
    \item \textbf{Prepending/Appending a Random String (PR/AR)} Inspired by \cite{somepalli2023diffusion}, we perturb the question by prepending or appending a short 6-digit random string to the question.
    \item \textbf{Extraction Blocking (EB)} For each original sample $(I,Q,a)$, we suggest adding to the training set a corresponding sample $(I^{-a},Q,\texttt{'ANSWER NOT PRESENT'})$. This approach is similar in spirit to the intuition behind the V-CSS part of the algorithm proposed in  \cite{chen2020counterfactual} to improve the grounding of VQA systems. 
\end{compactitem}

\input{figures/i_perturb/anls_results.tex}

\emph{Results}: We observe that although ITP and PR/AR can reduce the amount of extractable information, they also yield a substantial drop in ANLS on a held-out validation set. Therefore they can only be implemented as mitigation strategies if the practitioners are willing to pay a cost in terms of performance. On the other hand, we observe EB to be extremely effective, reducing to 0 the amount of extractable samples for most models.  Furthermore, although we apply the technique by augmenting the original training set using the context $(I^{-a}, Q)$, it is also generalizes to adversaries that query the model with the approaches considered in \cref{sec:ctx_importance} (see \cref{tab:eb_countext}), while producing an increase in the ANLS (in a similar way V-CSS does in \cite{chen2020counterfactual}).

%% file: figures/i_perturb/anls_results.tex
\begin{table}[]
    \begin{footnotesize}
    \centering
    \begin{tabular}{@{}l cccc@{}}
    \toprule 
    $\Delta$ ANLS / $|M|$ & PR & AR & ITP & EB (Ours) \\

    \midrule 
Donut & -3.4 / 38 & -3.1 / 34 & -12.5 / 26 & \textbf{+1.2 / 2}\\ 
Pix2Struct-B & -2.9 / 40 & -1.9 / 35 & -12.9 / 28 & \textbf{+3.4 / 0}\\
Pix2Struct-L & -2.6 / 37 & -2.0 / 33 & -13.8 / 25 & \textbf{+2.1 / 0}\\
PaLI-3 & -3.7 / 4 & -3.2 / 3 & -8.1 / 9 & \textbf{+1.5 / 0} \\
\bottomrule 
    \end{tabular}
    \end{footnotesize}
    \caption{Variation of ANLS (utility metric for DocVQA) and amount of extractable samples in $M$ for various countermeasures with respect to the standard training procedure.}
    \label{tab:anls_delta}
\end{table}

%% file: 7_conclusion.tex
\section{Conclusion} \label{section:conclusion}
In this study we have analysed the memorization abilities of three recent Document-Based VQA systems. We have shown these models can memorize information that is unique or sporadically repeated across the training set and it can be extracted when the model is prompted with incomplete context. We have introduced an extension of the Counterfactual Memorization and Simplicity scores that reveals that the memorized information identified by our attribution method is indeed also memorized according to these more computationally expensive scores. 
We have analysed the influence of the context on the extractability of samples, and studied the effectiveness of a few heuristic techniques, one of which results in a reduction of the amount of extractable samples and improves the test performance.

\section*{Impact statement} 
This paper shows it is possible for a malicious user to prompt a model to reveal training data. This phenomenon is studied in a worst-case but plausible condition in which the attacker knows the training image and question, except for the answer. Our study only represents a starting step in the direction of prompting VLMs to elicit the extraction of private data. It may be possible for an attacker to develop more sophisticated attack strategies. Such strategies can be used both in a beneficial way (e.g., for organizations to audit the privacy preserving properties of their systems) or maliciously (e.g., for an attacker to obtain confidential information).

 In this study we have used public data, and for further caution we have anonymised all the sensitive samples we reported in our qualitative analysis. Indeed, in some parts of the world the Right To Be Forgotten is in place, and the individuals whose data is reported in the considered public dataset my ask for their data to be cancelled. When performing our quantitative analysis, we report aggregate numbers and described the extractable samples without revealing their exact content for the same reasons. Therefore, we expect no individual or organization to be harmed by reporting our results. 

Furthermore, although we propose a countermeasure (EB) that is effective across all the attack scenarios we considered, it is still a heuristic approach and may not prevent extraction in case more sophisticated attack techniques are developed. Furthermore, it may hypothetically introduce a "side-channel" that an adversary might exploit to increase the exposure to membership inference attacks: if the model responds with the default negation, this may be seen as an index the sample was in the training set. This may not be relevant for several applications, where the information to be protected is not the membership of a document to the training set but the specific content of the document, but may be problematic in other applications. An obvious solution would be to apply Differentially Private (DP) training. Since DP provides guarantees about the likelihood of success of Membership Inference Attacks (MIA), but no closed form formula is available to translate the MIA guarantees into extraction prevention  guarantees, practitioners could consider tuning $(\epsilon,\delta)$ so as to empirically reduce extraction to zero. However, scaling DP to VLMs without causing significant utility degradation is a complex task that requires extensive and difficult parameter tuning \citep{kurakin2022training}, since noise addition and norm clipping could impact one of the two modalities disproportionately \citep{Hu2022M4IMM}. 
beyond the scope of this work.

\section*{Acknowledgements} We thank Chiyuan Zhang for his valuable feedback on our draft. We thank  Kenton Lee, Xi Chen and the PaLI team for their valuable technical support. We thank Michal Lukasik and Vaishnavh Nagarajan, for insightful discussions about memorization, our draft and how to frame our work in the context of the literature. We thank Samuele Marro for proofreading the work. The project was entirely funded by Google. Francesco Pinto's PhD is funded by the European Space Agency.

%% file: 8_appendix.tex
\section{Computational Cost of Training} \label{sec:computation_cost}

\paragraph{Donut} Fine-tuning Donut at maximum input resolution requires 64 A100 GPUs for a day.  Given its relatively compact size (176M parameters), Donut can be trained on  high-resolution input images (2560 $\times$ 1920 $\approx$ 5M pixels), a crucial aspect for achieving optimal performance. Lowering the resolution can significantly reduce the cost of training, however, as we observe, it increases the tendency of the model to memorize the training data and reduces the generalization capabilities of the models. Therefore it is not recommended.

\paragraph{Pix2Struct} Fine-tuning Pix2Struct Base, independently of the resolution, requires 32 TPUv2 for about 5 hours. Training Pix2Struct Large, independently of the resolution, requires 64 TPUv2 for about 5 hours. Due to its relatively larger size, the smaller model is fine-tuned at a resolution of about $1.2M$ pixels, while the larger model is fine-tuned at a resolution of about $0.8M$ pixels. 

\paragraph{PaLI-3} Fine-tuning PaLI-3 64 TPUv2 for 15 hours. Due to its size (5B parameters), it is typically fine-tuned at a resolution of approximately $1.1M$ pixels (1064 $\times$ 1064).  

\paragraph{Computing the memorization scores} The amount of compute needs to be multiplied by the number of runs for each measurement: for the simplest attribution method we consider, we only need 2 runs; for the counterfactual extractable memorization and simplicity scores, we need to perform 50 runs. Performing more is both computationally prohibitive and expensive for the storage of the largest models we consider. 

\section{Further results} 
\paragraph{Effectivenes of EB for prompting strategies not used in the training set} In \cref{sec:ctx_importance} we have considered several ways to prompt the model. Since EB includes only samples using a worst-case prompting strategy $(I^{-a},Q)$, it may be natural to wonder whether EB is still effective if an adversary prompts the model in different ways. We observe the technique is actually still extremely effective, see \cref{tab:eb_countext}

\input{figures/i_perturb/eb_context}

\section{PII categories and their frequencies in the canaries}\label{app:pii_docvqa}

\begin{figure}[t]
     \centering
     \includegraphics[width=0.5\linewidth]{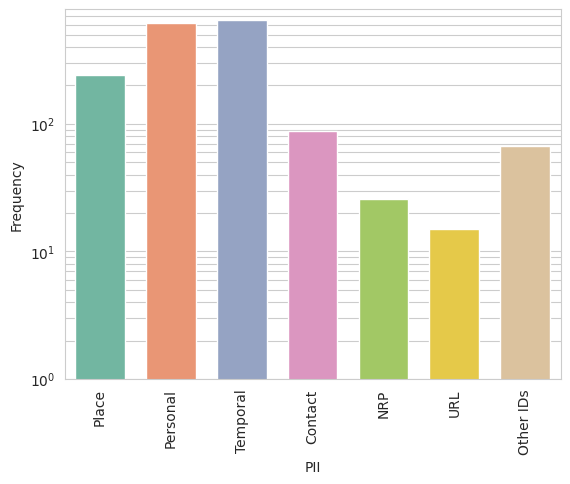}
     \caption{Frequency of different types of Personally Identifying Information (PII) in the canaries set $\mathcal{D}^C$.}
     \label{fig:pii_freq}
 \end{figure}

We manually annotate each answer in the canaries set as either PII or non-PII. We also classify each PII element as one of the following classes: Places, Person, Temporal, Contact (Phone/Fax/Email), NRP (Nationality Religion Politic), URL, and other forms of IDs (e.g. card numbers, serial numbers of tickets, document or people numerical identifiers etc.). The distribution of PII in the canaries set $\mathcal{D}^C$ is reported in \cref{fig:pii_freq}.

\section{Further Related Works} 
\label{sec:further_related_works}

\subsection{Document-Based Visual Question Answering}
\label{app:docvqa_models}
Given the greater simplicity of solving the VQA problem by separating the tasks of document reading and document understanding, OCR-reliant systems have been the state of the art for a few years \cite{tito2022hierarchical, huang2022layoutlmv3}. However, as argued by \cite{Donut}, OCR-reliant systems have the disadvantage of requiring an expensive OCR-preprocessing step, making the inference cost higher in case high-quality OCR results are required, with errors of the OCR system propagating to the VQA component. The phenomenon is particularly apparent for languages with complex character sets, requiring an expensive post-OCR correction module \cite{postocrcomponent,postocrcomponent2}. For these reasons, OCR-free systems like \cite{Donut, Pix2Struct} have received increasing attention, with state-of-the-art models like PALI-3 \cite{PALI3} closing the performance gap between the OCR-reliant and OCR-free models. In this work we mainly focus on three state-of-the-art OCR-free systems that differ in model size, architecture and pre-training stages. We consider both \textbf{Donut} \cite{Donut} and \textbf{Pix2Struct} \cite{Pix2Struct} among the set of models that are specialised to perform document understanding. We also consider \textbf{PALI-3} \cite{PALI3}, a foundational vision-language model that can be fine-tuned in order to solve the task of document understanding, achieving state-of-the-art performance. 

\subsection{Relations to Distributional Shortcut Learning in VQA} 
\label{sec:shortcuts} 

It is known that VQA systems can produce correct responses due to their ability to learn and leverage the frequent association of a specific answer to some question (linguistic shortcut) 
\cite{RevisitingVQA,  CounterfactualVQA,MakingVMatterVQA,chen2020counterfactual}. For instance, if the question is \emph{``What is the colour of the grass?"}, if the grass is green in most of the training images for which the question is asked, the model will respond green independently of the actual colour in the considered test image. This type of shortcuts does not need to be exclusively linguistic, and may involve the frequent co-occurrence of elements in the input image (visual shortcut) or their combination with specific words in the question (multimodal shortcut)  \cite{dancette2021beyond,Si2022LanguagePI}. In other terms, VQA systems can learn simple rules relying on spurious but predictive features that co-occurr across multiple samples in order to respond accurately even when the input image lacks the considered information or contradicts it. 

The concurrent work of \cite{tito2023privacy} has shown this phenomenon occurring also in document-based Visual Question Answering. The authors propose a new federated learning dataset containing invoices from several data providers. Since a provider's information (specifically, their name and email address) is \emph{repeated across several invoices} that share visual and linguistic similarities (e.g., identical layout, formatting, logos, fields etc.), a model can infer a provider's name or email address correctly on \emph{previously unseen test} documents from the known provider that do not contain the requested information. In contrast, we focus on \emph{centralised} training and perform attacks on training documents. Our analysis aims at factoring out the cases when models can extract information by leveraging knowledge learnt from other samples (which we consider as a form of generalization rather than memorization). While their goal is to protect the identity of providers (in a federated, group privacy setting), our goal is to protect the individual answers. 

%% file: figures/i_perturb/eb_context.tex
\begin{table}[]
    \centering
    \resizebox{\textwidth}{!}{\begin{tabular}{c|ccc|cccc|cccc}
    \toprule 
    $|M|$/ $\#$PII & No Text & Paraphrasing & Shuffling & R5° & R10° & T20px & T200px & B$\times$2 & B$\times$1.3 & B$\times$0.8 & B$\times$0.5 \\

    \midrule 
Donut & 0 / 0 & 0 / 0 & 0 / 0 & 1 / 0 & 0 / 0 & 0 / 0 & 0 / 0 & 6 / 1 & 6 / 1 & 2 / 0 & 5 / 0 \\ 
Pix2Struct-B  & 0 / 0 & 1 / 0 & 1 / 0 & 2 / 0 & 2 / 0 & 0 / 0 & 0 / 0 & 0 / 0 & 4 / 0 & 4 / 0 &  0 / 0 \\
Pix2Struct-L  & 1 / 0  & 0 / 0 & 0 / 0 & 1 / 0 & 1 / 0 & 0 / 0 & 0 / 0 & 0 / 0 & 4 / 0 & 2 / 0 & 0 / 0 \\
PaLI  & 0 / 0 & 0 / 0 & 2 / 0 & 1 / 0 & 1/ 0 & 0 / 0 & 2 / 0 & 3 / 0 & 1 / 0 & 0 / 0 & 1 / 0 \\
\bottomrule 
    \end{tabular}}
    \caption{Effectiveness of extraction blocking for the various contexts portrayed in \cref{fig:table2}. Notice, we do not include in the training sets any of the contexts we consider in this table. This indicates the protection offered by Extraction Blocking extends beyond the types of context provided at training time. }
    \label{tab:eb_countext}
\end{table}